\documentclass{article} 
\usepackage{iclr2027_conference,times}

\usepackage{amsmath,amsfonts,bm}

\def\eqref#1{equation~\ref{#1}}

\def\1{\bm{1}}

\DeclareMathAlphabet{\mathsfit}{\encodingdefault}{\sfdefault}{m}{sl}
\SetMathAlphabet{\mathsfit}{bold}{\encodingdefault}{\sfdefault}{bx}{n}

\usepackage{amsthm}

\usepackage{hyperref}
\usepackage{url}

\usepackage{graphicx}
\usepackage{booktabs}
\usepackage{adjustbox}
\usepackage{nicematrix}
\usepackage{siunitx}
\usepackage{multirow}
\usepackage{threeparttable}
\usepackage{enumitem}

\definecolor{light}{rgb}{0.5, 0.5, 0.5}

\definecolor{answerblue}{rgb}{0.21,0.37,0.57}

\usepackage[capitalize,nameinlink]{cleveref}
\crefname{figure}{Fig.\@}{Figs.\@}
\Crefname{figure}{Figure\@}{Figures\@}
\crefname{table}{Table\@}{Tables\@}
\Crefname{table}{Table\@}{Tables\@}

\usepackage{acronym}
\acrodef{asd}[ASD]{anomalous sound detection}
\acrodef{eat}[EAT]{efficient audio transformer}
\acrodef{knn}[kNN]{k-nearest neighbor}
\acrodef{ldn}[LDN]{local density-based score normalization}
\acrodef{lid}[LID]{local intrinsic dimensionality}
\acrodef{npd}[NPD]{normalized pseudo discrepancy}

\title{Can We Predict kNN-Based Anomalous Sound Detection Performance from Embedding-Space Geometry?}

\author{Kevin Wilkinghoff$~^{1,2}$, Zheng-Hua Tan$~^{1,2}$ \\
$^{1}$Department of Electronic Systems, Aalborg University, Denmark\\$^{2}$Pioneer Centre for Artificial Intelligence, Denmark\\
\texttt{kevin.wilkignhoff@ieee.org, zt@es.aau.dk} \\
}

\iclrfinalcopy 
\begin{document}

\maketitle

\begin{abstract}
Anomaly detection systems are often trained using normal data alone, while model selection and evaluation typically require labeled anomalies. We study anomaly-free selection for \ac{knn}-based anomaly detectors operating on embedding spaces, using pretrained audio representations and temporal pooling for first-shot industrial \acl{asd} as a practically relevant application. We derive a lower bound on the area under the ROC curve (AUC) at the score level and use it to analyze \ac{knn}-based detection, relating score variability to the geometry of the embedding space. Under a local scaling model, we characterize how density variation, intrinsic-dimensional heterogeneity, and cross-domain mismatch contribute to inlier-score variability, and show how \ac{ldn} affects these contributions. We then investigate anomaly-free model selection and show that inlier score variance alone does not reliably predict performance across different representations. To address this limitation, we introduce simple pseudo-anomaly probes that provide a reference for estimating relative score separation. Experiments on the DCASE 2022--2025 benchmarks, spanning four pretrained audio encoders and 208 candidate systems evaluate  both \ac{knn} and \ac{ldn}. The results show that \ac{ldn} improves detection performance, while pseudo-anomaly-based estimators substantially improve anomaly-free model selection. The AUC Bound provides robust selection across pseudo-anomaly constructions, and a simple metadata-free construction offers a competitive alternative to existing approaches, even outperforming conventional development-set selection when combined with \ac{ldn}.
\end{abstract}

\section{Introduction}
Anomalies are often unavailable for training because they are rare, diverse, and difficult to obtain. Consequently, many anomaly detection systems rely on normal data alone for training \citep{chandola2009anomaly,pang2022deep,ruff2018deep}. However, model selection, hyperparameter tuning, and evaluation typically rely on labeled anomalies to assess detection performance. This raises a fundamental question: Can anomaly detection performance be predicted without access to anomalous data? We investigate this question for commonly used distance-based detectors and examine whether the geometry of the embedding space, which governs the distances and neighborhood relationships upon which these detectors rely, contains sufficient information to predict detection performance.

In modern anomaly detection systems, the underlying geometry is induced by representations learned from the data \citep{han2025exploring,ruff2018deep,ruff2021unifying,pang2022deep}. 
Images \citep{roth2022towards}, audio signals \citep{wilkinghoff2021sub-cluster}, and time series \citep{hundman2018detecting} are commonly mapped into vector embeddings using pre-trained or task-specific models.
In this representation space, anomalies are identified through their deviations from the reference data, typically using distance- or density-based scoring functions.
Among these methods, \ac{knn} detectors remain a widely used class of distance-based detectors and are competitive across diverse anomaly detection benchmarks \citep{bukhsh2023ood,campos2016evaluation,sun2022ood}. They make few modeling assumptions, avoid parametric density estimation, and rely on local geometric relationships among data points, directly linking anomaly scores to local data geometry.
First-shot industrial \ac{asd} is a particularly relevant setting for this work, with pretrained audio representations and nearest-neighbor-based scoring achieving state-of-the-art performance on recent DCASE benchmarks \citep{saengthong2025deep,wilkinghoff2026temporal,saengthong2026sub,fujimura2026anomalous}.

However, the performance of \ac{knn}-based detectors can vary considerably across datasets, embeddings, and parameter choices \citep{campos2016evaluation}. Even when derived from the same normal data, different representations can yield very different detection performance. Although the geometry of normal data determines the scores produced by distance-based detectors, it remains unclear which geometric properties are predictive of detection performance and whether these signals are comparable across different representations.

In this work, we study anomaly-free model selection for \ac{knn}-based anomaly detectors operating on embedding spaces,
using pretrained audio representations and temporal pooling strategies as our experimental setting. Our main contributions are as follows:

\begin{itemize}[noitemsep, topsep=0pt]
\item We derive a lower bound on the area under the ROC curve (AUC) in terms of inlier and outlier score variances and mean separation. The bound provides a theoretical framework for analyzing detector reliability through the geometric properties of normal data.
\item We use this bound to analyze how density variation, intrinsic-dimensional heterogeneity, and cross-domain mismatch contribute to inlier-score variability. As an example, we show that \ac{ldn} removes the leading-order effect of density variation on inlier-score variance and suppresses the density-dependent component of cross-domain score mismatch.
\item We show that inlier-score variance alone is insufficient for anomaly-free model selection, while simple pseudo-anomaly probes provide the missing score-separation reference. The AUC Bound provides robust model selection across different pseudo-anomaly constructions. Moreover, a simple metadata-free Feature construction outperforms AutoUAD's Gaussian pseudo-anomaly baseline overall, and, when combined with \ac{ldn}, even surpasses conventional development-set selection on evaluation data.
\end{itemize}

\section{Related Work}
\label{sec:related_work}

Evaluation of anomaly detectors typically relies on ranking metrics such as AUC \citep{clemencon2008ranking}, which require labeled inlier and anomalous samples. Related work has addressed performance characterization in novelty detection through mixture proportion estimation and semi-supervised risk analysis \citep{blanchard2010semi}. Unsupervised model selection has been studied through internal evaluation criteria \citep{ma2023need}, which rely on heuristic measures of score separability, score distributions, or agreement between candidate models, meta-learning from historical labeled tasks \citep{zhao2021automatic}.
We take a complementary, detector-specific approach, deriving performance estimates from embedding-space geometry for \ac{knn}-based \ac{asd}.

Our geometric analysis builds on classical results linking \ac{knn} distances to local sampling density and intrinsic dimension \citep{fukunaga1973optimization,penrose2000central,singh2016finite}, as well as work using intrinsic-dimension estimates to characterize representation geometry and \ac{knn} behavior \citep{levina2004maximum,houle2013dimensionality,amsaleg2015estimating,ma2018characterizing}. Density variation has also motivated local normalization of distance-based anomaly scores, including recent \ac{ldn} approaches \citep{wilkinghoff2025local,matsumoto2025adjusting}. While these works characterize local geometry or normalize scores, they do not establish how these geometric properties relate quantitatively to detection performance or how they can support model selection under domain shifts. We build on these results to connect local scaling and intrinsic dimension to \ac{knn} score variability and derive a lower bound on detection performance.

Synthetic or auxiliary outliers are commonly used to train or regularize anomaly detectors and representation models \citep{hendrycks2019deep,chen2023effective,wilkinghoff2024why}. They have also been used for anomaly-free model selection: SWSA constructs synthetic validation anomalies from normal images \citep{fung2025model}, while AutoUAD introduces the \ac{npd}, which compares scores on held-out inliers with those of Gaussian samples generated from the training data \citep{dai2025autouad}. Despite inducing identical candidate rankings on the same inlier and pseudo-outlier scores, \ac{npd} and our AUC Bound differ in their theoretical foundations: \ac{npd} is a normalized score discrepancy analyzed as a surrogate for expected test performance, whereas our bound follows directly from an inequality on AUC, providing an explicit lower bound on detection performance. Accordingly, our contribution lies in grounding performance estimation in \ac{knn} score geometry and systematically investigating the effects of pseudo-outlier construction, domain shifts, and local density normalization on model selection.

\section{Geometric Analysis of \acs{knn} Reliability}
\label{sec:analysis}

Anomaly detection performance depends on the distributions of inlier and outlier scores. In anomaly-free settings, the outlier distribution is unavailable, making the full score distributions difficult to characterize. We therefore derive an AUC lower bound in terms of the first two moments of the score distributions, allowing us to analyze how score variability affects detection performance.

\subsection{A Variance-Based Lower Bound on the AUC}
\label{sec:single_domain}

Let $z_{\mathrm{in}}$ and $z_{\mathrm{out}}$ denote the random variables corresponding to the log-scores assigned to inlier and outlier samples, respectively. We consider independently drawn inlier and outlier samples, with both scores computed against the same fixed reference set. We treat the reference set as fixed throughout, so all expectations and variances are understood conditionally on it. Thus, $z_{\mathrm{in}}$ and $z_{\mathrm{out}}$ are independent. Define the score difference as
\[
\Delta = z_{\mathrm{out}} - z_{\mathrm{in}}.
\]
By definition, the AUC can be expressed as the probability that a randomly drawn outlier receives a higher score than a randomly drawn inlier \citep{clemencon2008ranking},
\[
\mathrm{AUC} = \mathbb{P}(\Delta > 0).
\]
Assume that \(\Delta\) has a finite second moment and positive mean, $\mu_\Delta = \mathbb{E}[\Delta] > 0$, corresponding to the conventional anomaly-score orientation in which higher scores indicate stronger evidence of anomalousness. Applying Cantelli's inequality \citep{ion2023sharp} to $\Delta$, together with the independence of the inlier and outlier scores, yields
\begin{equation}
\label{eq:auc_variance_bound}
\mathrm{AUC}
\ge
\frac{\mu_\Delta^2}
{
\mathrm{Var}(z_{\mathrm{out}})
+
\mathrm{Var}(z_{\mathrm{in}})
+
\mu_\Delta^2}.
\end{equation}
The AUC Bound is closely related to AutoUAD's \ac{npd}, which normalizes squared mean score separation by twice the sum of score variances. We derive it directly from the AUC using Cantelli's inequality. A dependence-robust version is given in \Cref{sec:auc_bound_proof}.

\begin{figure}[t]
\centering
\begin{adjustbox}{max width=0.47\columnwidth}
\includegraphics[width=\linewidth]{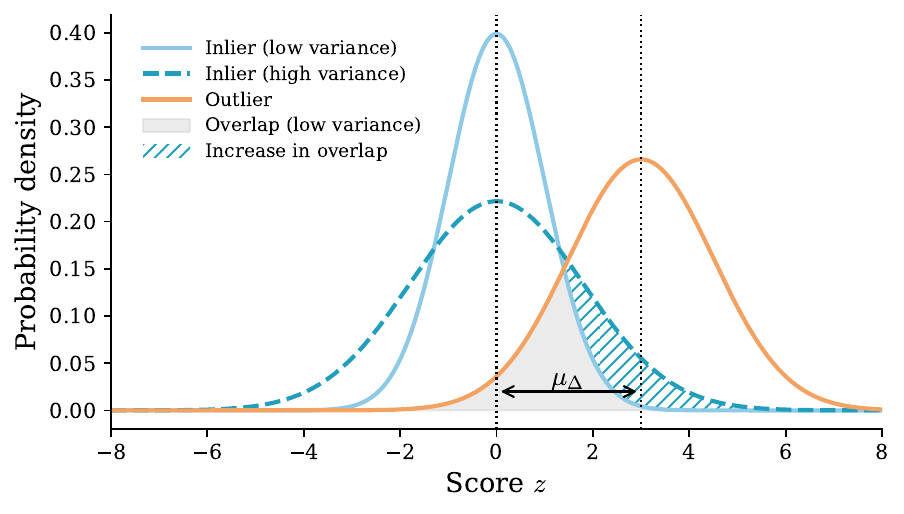}
\end{adjustbox}

\caption{
Effect of inlier score variance on ranking reliability. For fixed mean separation $\mu_\Delta$ and outlier variance, increasing inlier variance increases score overlap. The light shaded region shows baseline overlap, while the hatched region shows additional overlap from variance inflation.
}
\label{fig:auc_bound}

\end{figure}

The AUC lower bound in \cref{eq:auc_variance_bound} increases with mean separation between inlier and outlier scores and decreases with their variances. The bound is sharp given only the first two moments of the score difference, with equality attained for a two-point distribution \citep{ion2023sharp}. In practice, the bound may be loose, but it still reflects how score separation and variance affect detection performance. In the anomaly-free setting, however, only the inlier-score distribution is observable. Therefore, the bound reveals both the information available from normal data and the information that is missing: While inlier-score variability can be estimated directly, the outlier variance and mean separation remain unknown. The effect of inlier variance under a fixed mean separation is illustrated in \cref{fig:auc_bound}.

\subsection{Geometric Decomposition of Inlier Score Variance}
\label{sec:geometric_decomposition}
While $\mathrm{Var}(z_{\mathrm{in}})$ is directly observable, it aggregates several geometric effects. To disentangle these contributions, we draw on local scaling theory to characterize how density variation, intrinsic-dimensional heterogeneity, and domain mismatch contribute to inlier score variability.

\subsubsection{Local scaling model}

Let $\mathcal{R}=\{x_1,\ldots,x_n\}\subset\mathcal{X}$ denote a
reference set of inlier samples in a metric space $(\mathcal{X},d)$.
For $x\in\mathcal{X}$, let $d_k(x)$ denote the distance from $x$
to its $k$-th nearest neighbor in $\mathcal{R}$, and let
\[
B(x,r)=\{y\in\mathcal{X}: d(x,y)\le r\}
\]
denote the metric ball of radius $r$ centered at $x$.

At sufficiently small scales, the inlier distribution behaves
locally like a low-dimensional measure, so that the probability
mass of a small ball grows approximately as a power of its
radius under standard intrinsic-dimension assumptions and
classical \ac{knn} scaling results
\citep{levina2004maximum,penrose2000central,singh2016finite}. Classical analyses typically assume a globally fixed intrinsic dimension. In practice, embedding spaces often exhibit
heterogeneous local geometry, motivating the use of a pointwise
intrinsic-dimension function $m(x)$ as in the \ac{lid} literature
\citep{houle2013dimensionality,amsaleg2015estimating}.

Formally, this local behavior can be described through the
following small-ball expansion commonly used in
intrinsic-dimension analysis:
\[
\mathbb{P}(B(x,r))
=
\rho(x)\, V_{m(x)}\, r^{m(x)}
\bigl(1 + \varepsilon(r)\bigr),
\]
for sufficiently small $r$, where $V_{m(x)}$ denotes the volume
of the unit ball in $\mathbb{R}^{m(x)}$ and
$\varepsilon(r)\to 0$ as $r\to 0$. The functions $m(x)$ and
$\rho(x)$ denote the local intrinsic dimension and density,
respectively.

As $n \to \infty$, the $k$-nearest neighbor radius $d_k(x)$ converges
to zero for fixed $k$, so the small-ball expansion applies at
$r=d_k(x)$. Classical results on \ac{knn} radii further imply that the
corresponding scaled probability mass
\[
U_{n,k}(x)
=
\frac{n}{k}\mathbb{P}(B(x,d_k(x)))
\]
converges in distribution to a non-degenerate random variable, rather
than to a constant \citep{fukunaga1973optimization,penrose2000central}.
Applying the small-ball expansion to this relation gives
\begin{equation}
\label{eq:dk_expansion}
d_k(x)
=
\left(
\frac{k\,U_{n,k}(x)}
{n\,\rho(x)\,V_{m(x)}}
\right)^{1/m(x)}
\bigl(1+\varepsilon_n(x)\bigr),
\end{equation}
where $\varepsilon_n(x)\to0$ as $n\to\infty$. Using
$z(x)=\log d_k(x)$, we obtain
\begin{equation}
\label{eq:log_score_expansion}
z(x)
=
\frac{1}{m(x)}
\left(
\log k-\log(n\rho(x)V_{m(x)})
\right)
+
\frac{1}{m(x)}\log U_{n,k}(x)
+
\epsilon_n(x)
\end{equation}
with $\epsilon_n(x)\to0$ as $n\to\infty$.
The stochastic term $\log U_{n,k}(x)$ does not vanish asymptotically
and therefore contributes irreducible \ac{knn} sampling noise to the
score variance. The leading term in \cref{eq:log_score_expansion}
instead captures the geometric component of score variability, which
depends on the local density $\rho(x)$ and intrinsic dimension
$m(x)$. The density contribution enters through $\log\rho(x)$, while
the intrinsic dimension affects both the reciprocal factor $1/m(x)$
and the volume term $V_{m(x)}$. Accordingly, the geometric component
captures density heterogeneity, intrinsic-dimension heterogeneity,
and their interaction.

\subsubsection{Variance Decomposition Under Domain Shift}
\label{sec:domain_shift}

The preceding analysis characterizes the sources of score variability within a domain. Under domain shift, this variability is further affected by differences in score statistics between domains. We therefore decompose the score variances and mean separation into their source and target domains.

The total score variance can be decomposed into within-domain variability
and a between-domain mismatch term. Let $P_S$ and $P_T$ denote source
and target-domain distributions, respectively, and let $\lambda$ denote the mixture
weight of the source domain. Then,
\begin{equation}
\label{eq:mixture_variance}
\begin{aligned}
\mathrm{Var}(z_{\mathrm{in}})
&=
\lambda\,\mathrm{Var}_{P_S}(z_{\mathrm{in}})
+
(1-\lambda)\,\mathrm{Var}_{P_T}(z_{\mathrm{in}})+
\lambda(1-\lambda)
\left(
\mathbb{E}_{P_S}[z_{\mathrm{in}}]
-
\mathbb{E}_{P_T}[z_{\mathrm{in}}]
\right)^2.
\end{aligned}
\end{equation}

The first two terms describe within-domain score variability, while the
last term captures the mismatch between the average score levels of the
two domains.
The same decomposition applies to the outlier score variance,
\begin{equation}
\mathrm{Var}(z_{\mathrm{out}})
=
\lambda\,\mathrm{Var}_{P_S}(z_{\mathrm{out}})
+
(1-\lambda)\,\mathrm{Var}_{P_T}(z_{\mathrm{out}})
+
\lambda(1-\lambda)
\left(
\mathbb{E}_{P_S}[z_{\mathrm{out}}]
-
\mathbb{E}_{P_T}[z_{\mathrm{out}}]
\right)^2.
\end{equation}

Similarly, the mean score separation can be written as
\begin{equation}
\label{eq:mixture_mismatch}
\begin{aligned}
\mu_\Delta
&=
\lambda\left(
\mathbb{E}_{P_S}[z_{\mathrm{out}}]
-
\mathbb{E}_{P_S}[z_{\mathrm{in}}]
\right)+
(1-\lambda)\left(
\mathbb{E}_{P_T}[z_{\mathrm{out}}]
-
\mathbb{E}_{P_T}[z_{\mathrm{in}}]
\right).
\end{aligned}
\end{equation}

These decompositions show that domain shift affects the AUC bound through both within-domain score variability and between-domain mismatch. When the domains have identical score distributions, the mismatch terms vanish.

\subsection{Case Study: Local Density Normalization}
\label{sec:ldn_analysis}

The AUC bound can also be used to analyze how an existing anomaly detection method changes score variability. We illustrate this with \ac{ldn} \citep{wilkinghoff2025local}, which normalizes the \ac{knn} distance by a local reference distance scale. Combining the bound with the local scaling model allows us to identify which components of score variability are removed by the normalization.

To isolate the geometric effect of \ac{ldn}, we temporarily ignore the
non-vanishing \ac{knn} sampling variability represented by the
$U_{n,j}(x)$ terms in the local scaling expansion.
Let
\[
\bar d_K(x)
=
\frac{1}{K}\sum_{j=1}^{K} d_j(x)
\]
denote the mean distance from $x$ to its $K$ nearest neighbors in the
reference set. Under the local scaling model,
\[
d_j(x)
=
\left(
\frac{j}{n\rho(x)V_{m(x)}}
\right)^{1/m(x)}
(1+\varepsilon_n(x,j)).
\]
Consequently,
\[
\bar d_K(x)
=
\left(
\frac{1}{n\rho(x)V_{m(x)}}
\right)^{1/m(x)}
C_K(m(x))
(1+\varepsilon_n(x,K)),
\]
where
$
C_K(m)
=
\frac{1}{K}\sum_{j=1}^{K}j^{1/m}
$
depends only on the intrinsic dimension $m$.

Substituting the local scaling expressions yields the locally density-normalized score 
\begin{equation}
\label{eq:ldn_score_expansion}
\tilde z(x)
=
\log d_k(x)-\log\bar d_K(x)
=
\frac{\log k}{m(x)}-\log C_K(m(x))
+
\tilde\varepsilon_n(x,k,K),
\end{equation}
where $\tilde\varepsilon_n(x,k,K)$ collects the remainder terms.
Importantly, the density $\rho(x)$ and the reference set size $n$ no
longer appear in the leading term. Thus, \ac{ldn} removes the
leading-order contribution of density variation and cancels the
$-\log n/m(x)$ term. The latter is responsible for a $O((\log n)^2)$ variance growth when intrinsic dimension varies across the data manifold, so both effects reduce inlier score variance and tighten the AUC bound.

The same cancellation applies to between-domain score mismatch.
Consider the source and target domains $P_S$ and $P_T$ in
\cref{sec:domain_shift}, and let $\tilde z_{\mathrm{in}}$ denote the
normalized score of an inlier. Assume that the source and target distributions have the same intrinsic dimension $m$, but may exhibit different local sampling densities $\rho_S(x)$ and $\rho_T(x)$. Then, the leading-order
term in \cref{eq:ldn_score_expansion} is identical in both domains and cancels out.
Therefore, the density-dependent component of the leading-order between-domain mismatch in \cref{eq:mixture_variance} vanishes, leaving only the finite-sample remainder:
\begin{equation}
\label{eq:ldn_domain_mismatch}
\mathbb{E}_{P_S}[\tilde z_{\mathrm{in}}]
-
\mathbb{E}_{P_T}[\tilde z_{\mathrm{in}}]
=
\mathbb{E}_{P_S}[\tilde\varepsilon_n]
-
\mathbb{E}_{P_T}[\tilde\varepsilon_n].
\end{equation}

The analysis explains the effect of \ac{ldn}: Within a domain, it removes the leading-order contribution of local density variation to score variability, while across domains it removes the corresponding density-dependent component of score mismatch. When the source and target domains have different intrinsic dimensions, a residual mismatch remains. The same bound relates score-variance minimization and cross-domain score normalization to embedding-space geometry: Recent domain-robust methods minimize score variance \citep{matsumoto2025adjusting}, while post-hoc normalization addresses cross-domain score mismatch \citep{saengthong2025deep}.

\section{Pseudo-Anomaly-Based Performance Estimation}
\label{sec:pseudo_outlier}

The geometric analysis shows that normal-data score statistics contain information about detection performance, even though the outlier score distribution remains unknown. Lower inlier score variance strengthens the AUC bound, motivating the inverse variance, $1/\operatorname{Var}(z_{\mathrm{in}})$, as an anomaly-free performance estimator. However, its representation-dependent scale limits comparability across candidate systems, while the unknown outlier distribution prevents direct evaluation of score separation.

To address this limitation, we construct \emph{pseudo-anomalies} directly in the embedding space and use their scores as proxies for the unknown outlier score distribution. This follows the general principle of existing approaches such as SWSA \citep{fung2025model}, which ranks candidates by AUC using image-specific synthetic anomalies, and AutoUAD \citep{dai2025autouad}, which uses Gaussian pseudo-validation samples for the \ac{npd}. By constructing pseudo-anomalies directly in the embedding space, our approach avoids the need to generate synthetic anomalous audio in the input domain and applies to \ac{knn}-based detectors across modalities and embedding models.

Given the resulting pseudo-outlier scores, defined as the distance to the nearest ($k=1$) reference sample, we consider two estimators that require only normal reference data and pseudo anomalies:

\begin{itemize}[noitemsep, topsep=0pt]
    \item \textbf{Pseudo AUC} computes the AUC using inlier and pseudo-outlier scores.

    \item \textbf{AUC Bound} replaces the unknown outlier moments in \cref{eq:auc_variance_bound} with pseudo-outlier moments. On identical inlier and pseudo-outlier scores, it yields the same candidate ranking as AutoUAD's \ac{npd}. We set the bound to zero whenever the positive-mean-difference assumption is violated ($\mu_\Delta\leq0$), thereby retaining a valid (trivial) lower bound in all cases.
\end{itemize}

\subsection{Pseudo-Anomaly Constructions}
\label{sec:pseudo_anomalies}

We investigate several constructions that introduce different deviations from normal embeddings.

\begin{itemize}[noitemsep, topsep=0pt]

\item \textbf{AutoUAD:}
Following \citet{dai2025autouad}, the reference embeddings are randomly split into training and validation subsets (70/30). A factorized Gaussian is fitted to the training embeddings and sampled to generate as many pseudo-anomalies as validation samples. Both are scored against the training subset.

\item \textbf{Random:}
Pseudo-anomalies are sampled independently from the isotropic standard normal distribution $\mathcal{N}(0,I)$.

\item \textbf{Feature:}
Each feature dimension is sampled with replacement from a randomly selected reference embedding, independently across dimensions.

\item \textbf{Sequence:}
Pseudo-anomalies are constructed by stitching embeddings from different samples prior to sequence pooling.

\item \textbf{Element:}
Individual elements of embedding sequences are randomly exchanged between different samples prior to sequence pooling.

\item \textbf{Cross-Domain, Cross-Class, Cross-Attribute:}
Pseudo-anomalies are reference embeddings from a different domain, semantic class, or semantic attribute, respectively.

\item \textbf{Diverse:}
Combines pseudo-anomalies from Feature, Cross-Domain, Cross-Class, and Cross-Attribute constructions.

\end{itemize}

Random, Feature, Sequence, and Element generate pseudo-anomalies from frame-level embeddings before temporal pooling to reduce pooling-specific biases. Cross-Domain, Cross-Class, and Cross-Attribute randomly subsample eligible reference embeddings to obtain as many pseudo-anomalies as inliers, keeping their counts comparable across constructions. A practical limitation of these constructions and Diverse is their reliance on metadata, which may not always be available. AutoUAD generates Gaussian pseudo-anomalies from pooled reference embeddings.

\section{Experimental Setup}
\label{sec:setup}

\subsection{Datasets}

Although the proposed approach is not specific to a sensing modality, we use the DCASE benchmark series as our application domain. It provides a large number of independent model-selection tasks spanning diverse monitored objects, operating conditions, and domain shifts.
Our evaluation covers the DCASE~2022--2025 benchmark datasets \citep{dohi2022description,dohi2023description,nishida2024description,nishida2025description}. Collectively, they comprise the MIMII-DG \citep{dohi2022mimiidg}, ToyADMOS2 \citep{harada2021toyadmos2}, ToyADMOS2+ \citep{harada2023toyadmos2+}, ToyADMOS2\# \citep{niizumi2024toyadmos2sharp}, ToyADMOS2025 \citep{harada2025toyadmos2025}, and IMAD-DS \citep{albertini2024imadds} datasets.

\par

Each benchmark comprises multiple independent machine-type-specific tasks, split into separate development and evaluation sets. Across all four benchmarks, this results in 84 development tasks and 90 evaluation tasks, yielding 174 independent model-selection tasks in total. In every task, only normal operation recordings are provided as reference data, whereas the corresponding test set contains both normal and anomalous recordings. The benchmarks evaluate domain generalization, with each reference set containing 990 source-domain and 10 target-domain recordings. Source and target domains are balanced during testing, but domain labels are not provided.

\subsection{Candidate Systems}

A meaningful evaluation of anomaly-free model selection requires a large and diverse set of candidate systems. To this end, we combine multiple self-supervised audio embedding models with a broad range of sequence pooling strategies, yielding 208 candidate systems in total.
Specifically, we consider OpenL3 \citep{cramer2019look}, BEATs \citep{chen2023beats}, \ac{eat} \citep{chen2024eat}, and Dasheng \citep{dinkel2024dasheng}. All models are used in their pre-trained form without task-specific fine-tuning. Exact configurations are given in \cref{appendix:model_configs}. For each model, we evaluate 52 pooling configurations: mean, max, generalized mean pooling (GeM) \citep{radenovic2019fine-tuning} with $p=1,\ldots,25$, and relative deviation pooling (RDP) \citep{wilkinghoff2026temporal} with $\gamma=1,\ldots,25$. Across the four models, this yields 208 candidate systems, evaluated with logarithmic nearest-neighbor distance ($k=1$) and \ac{ldn} ($K=2$).
We evaluate all pseudo-anomaly constructions described in \cref{sec:pseudo_anomalies} for candidate system selection, using machine type and operating condition as the class and attribute labels for Cross-Class and Cross-Attribute, respectively.

\subsection{Evaluation Metrics}

The proposed estimators produce proxy scores for ranking candidate embedding models. We evaluate their quality from three perspectives. Global ranking performance is measured by the Spearman rank correlation, $\rho_{\mathrm{rank}}\in[-1,1]$, with undefined correlations caused by constant predictions set to zero. Local ranking is measured by pairwise ranking accuracy, $A_{\mathrm{pair}}\in[0,1]$, defined as the fraction of correctly ordered model pairs. Finally, model selection performance is measured by the regret
\[
R=\mathrm{AUC}_{\mathrm{best}}-\mathrm{AUC}_{\mathrm{selected}},
\]
which quantifies the loss from selecting the predicted best model rather than the optimal model. Thus, $R=0$ corresponds to perfect selection. When multiple candidates attain the same best estimator value, we define the selected AUC as the mean AUC across all tied candidates. This corresponds to the expected selected AUC under uniform random tie-breaking.
Unless stated otherwise, all metrics are computed independently for each section and averaged across all sections and datasets.
All reported confidence intervals (CIs) are 95\% bootstrap intervals. Statistical significance is assessed using paired CIs of differences in selected AUCs across all 174 model-selection tasks (90 evaluation tasks for comparisons involving Fixed Selection).

\section{Experimental Results}
\begin{table*}[t]
\centering
\caption{
Anomaly-free model selection on the DCASE~2022--2025 datasets.
Entries report the Spearman rank correlation ($\rho_{\mathrm{rank}}$), pairwise ranking accuracy ($A_{\mathrm{pair}}$), and regret ($R$, in percent).
The \emph{Overall} columns report the mean across both scoring functions and dataset splits.
Boldface and underlining indicate the best overall and metadata-free pseudo-anomaly constructions, respectively.
}
\label{tab:anomaly_free_model_selection}

\setlength{\tabcolsep}{3pt}
\begin{threeparttable}
\begin{adjustbox}{max width=\textwidth}
\begin{NiceTabular}{l*{16}{c}}
\toprule

&
&
&
&
&
\multicolumn{6}{c}{\ac{knn}}
&
\multicolumn{6}{c}{\ac{ldn}}
\\

\cmidrule(lr){6-11}
\cmidrule(lr){12-17}

&
&
\multicolumn{3}{c}{Overall}
&
\multicolumn{3}{c}{Development}
&
\multicolumn{3}{c}{Evaluation}
&
\multicolumn{3}{c}{Development}
&
\multicolumn{3}{c}{Evaluation}
\\

\cmidrule(lr){3-5}
\cmidrule(lr){6-8}
\cmidrule(lr){9-11}
\cmidrule(lr){12-14}
\cmidrule(lr){15-17}

Estimator
&
Pseudo-Anomaly
&
$\rho_{\mathrm{rank}}\uparrow$
&
$A_{\mathrm{pair}}\uparrow$
&
$R\downarrow$
&
$\rho_{\mathrm{rank}}\uparrow$
&
$A_{\mathrm{pair}}\uparrow$
&
$R\downarrow$
&
$\rho_{\mathrm{rank}}\uparrow$
&
$A_{\mathrm{pair}}\uparrow$
&
$R\downarrow$
&
$\rho_{\mathrm{rank}}\uparrow$
&
$A_{\mathrm{pair}}\uparrow$
&
$R\downarrow$
&
$\rho_{\mathrm{rank}}\uparrow$
&
$A_{\mathrm{pair}}\uparrow$
&
$R\downarrow$
\\

\midrule

Oracle
&--
&$1.00$&$1.00$&$0.00$
&$1.00$&$1.00$&$0.00$
&$1.00$&$1.00$&$0.00$
&$1.00$&$1.00$&$0.00$
&$1.00$&$1.00$&$0.00$
\\

Random Selection
&--
&$0.00$&$0.50$&$7.93$
&$0.00$&$0.50$&$7.79$
&$0.00$&$0.50$&$7.99$
&$0.00$&$0.50$&$7.92$
&$0.00$&$0.50$&$8.01$
\\

Fixed Selection\tnote{a}
&--
&--&--&$4.35$
&--&--&$3.56$
&--&--&$4.51$
&--&--&$4.56$
&--&--&$4.76$
\\

\midrule

Inlier Variance
&--
&$0.16$&$0.56$&$5.53$
&$0.13$&$0.54$&$6.07$
&$0.17$&$0.56$&$5.30$
&$0.17$&$0.56$&$5.57$
&$0.18$&$0.56$&$5.19$
\\

\midrule

\multirow{9}{*}{AUC Bound}

&AutoUAD
&$0.24$&$0.59$&$5.67$
&$0.19$&$0.57$&$6.82$
&$0.35$&$0.63$&\underline{$4.73$}
&$0.11$&$0.54$&$6.19$
&$0.29$&$0.60$&$4.95$
\\

& Random
&$0.08$&$0.53$&$5.36$
&$0.06$&$0.52$&\underline{$5.48$}
&$0.08$&$0.54$&$5.42$
&$0.11$&$0.54$&$5.07$
&$0.06$&$0.53$&$5.48$
\\
& Sequence
&$0.01$&$0.50$&$8.62$
&$-0.04$&$0.48$&$8.44$
&$0.00$&$0.50$&$9.03$
&$-0.02$&$0.49$&$9.60$
&$0.08$&$0.52$&$7.41$
\\
& Feature
&\underline{$0.31$}&\underline{$0.62$}&\underline{$4.97$}
&$0.22$&$0.59$&$5.99$
&\underline{$0.38$}&\underline{$0.64$}&$5.17$
&\underline{$0.21$}&\underline{$0.58$}&\underline{$5.06$}
&\underline{$0.41$}&\underline{$0.65$}&\underline{$3.67$}
\\
& Element
&$0.04$&$0.51$&$8.30$
&$-0.05$&$0.48$&$8.47$
&$0.10$&$0.53$&$8.38$
&$-0.04$&$0.49$&$8.34$
&$0.13$&$0.54$&$7.99$
\\
& Cross-Domain\tnote{b}
&$0.22$&$0.59$&$5.76$
&$0.29$&$0.61$&$5.89$
&$0.33$&$0.63$&$4.42$
&$0.17$&$0.56$&$5.72$
&$0.09$&$0.54$&$7.02$
\\
& Cross-Class\tnote{b}
&$0.28$&$0.60$&$4.98$
&$0.26$&$0.60$&$4.50$
&$0.36$&$0.63$&$4.14$
&$0.25$&$0.59$&$5.66$
&$0.26$&$0.59$&$5.60$
\\
& Cross-Attribute\tnote{b}
&\pmb{$0.36$}&\pmb{$0.64$}&$4.41$
&$0.33$&\pmb{$0.63$}&$4.46$
&$0.37$&$0.64$&\pmb{$3.29$}
&\pmb{$0.35$}&\pmb{$0.62$}&$6.15$
&$0.40$&$0.65$&$3.75$
\\
& Diverse\tnote{b}
&$0.31$&$0.61$&$4.33$
&$0.32$&$0.62$&\pmb{$3.96$}
&$0.37$&$0.63$&$3.78$
&$0.26$&$0.59$&$5.09$
&$0.29$&$0.60$&$4.48$
\\

\midrule

\multirow{9}{*}{Pseudo AUC}
&AutoUAD
&$0.19$&$0.57$&$6.29$
&$0.17$&$0.56$&$6.48$
&$0.27$&$0.59$&$5.89$
&$0.08$&$0.53$&$6.79$
&$0.24$&$0.58$&$6.01$
\\

& Random
&$0.02$&$0.50$&$7.89$
&$0.00$&$0.50$&$7.80$
&$0.01$&$0.50$&$7.97$
&$0.02$&$0.50$&$7.91$
&$0.06$&$0.51$&$7.89$
\\
& Sequence
&$-0.02$&$0.49$&$8.47$
&$-0.07$&$0.47$&$8.83$
&$-0.01$&$0.49$&$9.02$
&$-0.06$&$0.48$&$8.82$
&$0.08$&$0.52$&$7.21$
\\
& Feature
&$0.28$&$0.60$&$5.54$
&\underline{$0.28$}&\underline{$0.60$}&$5.49$
&$0.35$&$0.63$&$5.29$
&$0.18$&$0.57$&$5.61$
&$0.31$&$0.61$&$5.78$
\\
& Element
&$0.02$&$0.51$&$8.33$
&$-0.06$&$0.48$&$8.19$
&$0.08$&$0.52$&$9.08$
&$-0.04$&$0.49$&$8.39$
&$0.10$&$0.53$&$7.65$
\\
& Cross-Domain\tnote{b}
&$0.16$&$0.56$&$6.69$
&$0.28$&$0.60$&$6.41$
&$0.26$&$0.60$&$5.90$
&$0.01$&$0.51$&$7.93$
&$0.07$&$0.53$&$6.53$
\\
& Cross-Class\tnote{b}
&$0.26$&$0.60$&$5.31$
&$0.26$&$0.60$&$4.92$
&$0.29$&$0.60$&$5.28$
&$0.25$&$0.59$&$5.41$
&$0.24$&$0.59$&$5.64$
\\
& Cross-Attribute\tnote{b}
&$0.35$&$0.63$&$4.80$
&\pmb{$0.34$}&\pmb{$0.63$}&$4.87$
&$0.33$&$0.63$&$4.23$
&$0.30$&$0.61$&$6.46$
&\pmb{$0.42$}&\pmb{$0.66$}&\pmb{$3.64$}
\\
& Diverse\tnote{b}
&\pmb{$0.36$}&$0.63$&\pmb{$4.28$}
&$0.32$&$0.62$&$4.52$
&\pmb{$0.42$}&\pmb{$0.65$}&$4.21$
&$0.31$&$0.61$&\pmb{$4.69$}
&$0.40$&$0.65$&$3.71$
\\

\bottomrule
\end{NiceTabular}
\end{adjustbox}
\begin{tablenotes}\footnotesize
    \item [a] Fixed Selection selects, for each DCASE dataset, the candidate system with the highest average ground-truth AUC on the development set and applies it unchanged to the corresponding evaluation set.
    \item [b] These pseudo-anomaly constructions require metadata labels for the reference samples.
    \end{tablenotes}
    \end{threeparttable}
\end{table*}

\Cref{tab:anomaly_free_model_selection} summarizes the model selection performance of all considered estimators. As expected, the Fixed Selection baseline substantially outperforms Random Selection (+3.58 pp [+2.88, +4.26]), demonstrating that anomaly detection performance is sufficiently consistent across related monitored objects to enable effective model selection when anomaly labels are available.

\subsection{Can anomaly detection performance be predicted without anomalies?}
\begin{figure}[t]
\centering
\begin{adjustbox}{max width=\columnwidth}
\includegraphics[width=\linewidth]{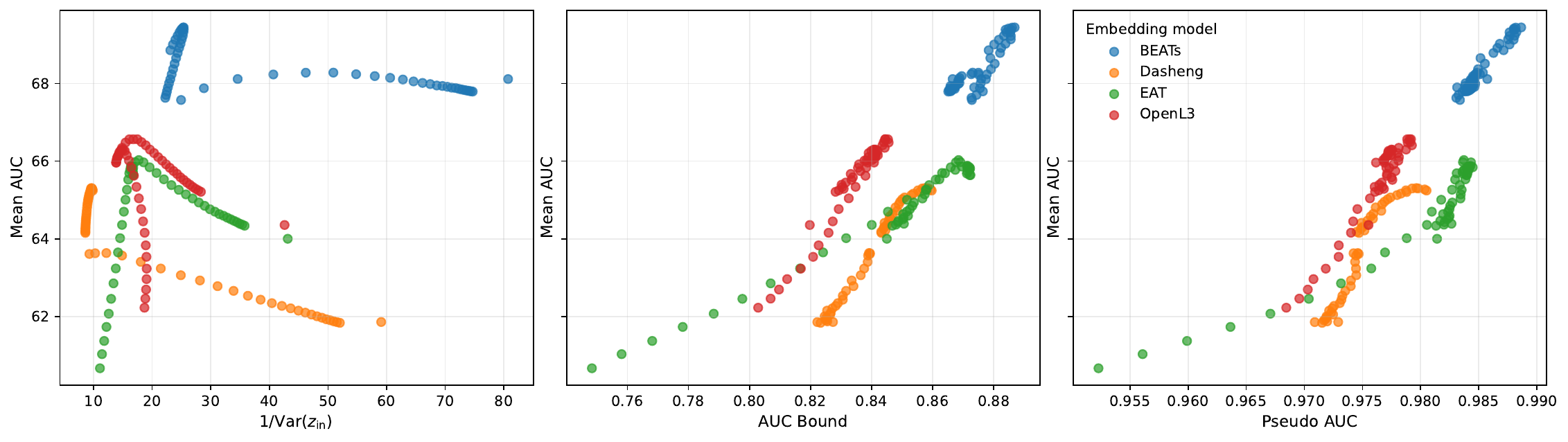}
\end{adjustbox}

\caption{
Relationship between anomaly-free performance estimators and mean AUC across the 208 candidate systems using \acs{knn} scoring. Each point represents one candidate system, averaged across all dataset splits. Within each embedding model, candidate systems form clear trajectories as the pooling strategy and its parameterization vary. The estimator-AUC relationship is strongly representation-dependent for inlier score variance, whereas this dependence is substantially reduced for the AUC Bound and Pseudo AUC estimators using Diverse pseudo-anomalies.
}
\label{fig:scatter}

\end{figure}

The simplest anomaly-free estimator, based solely on inlier score variance, already outperforms Random Selection (+2.40 pp [+1.73, +3.07]), demonstrating that normal data contain useful information for model selection. However, it remains inferior to Fixed Selection (-1.18 pp [-1.85, -0.50]). \Cref{fig:scatter} illustrates why this signal is difficult to compare across representations: Candidate systems form distinct embedding-specific trajectories as the pooling strategy and its parameterization vary, with substantially different score-variance scales.

In contrast, the AUC Bound and Pseudo AUC show more consistent estimator–AUC relationships across embedding models (\cref{fig:scatter}). Pseudo-anomalies provide a reference for score separation that is unavailable from the inlier distribution alone. Importantly, this benefit is not limited to a particular pseudo-anomaly construction: a range of constructions improves model selection over inlier variance alone, with the best-performing approaches even surpassing Fixed Selection on the evaluation sets, despite requiring no anomalous validation samples.

Overall, these results show that anomaly detection systems can indeed be ranked effectively without anomalous validation samples. However, the large performance differences across pseudo-anomaly constructions indicate that their effectiveness depends strongly on how they are generated. We therefore next investigate which pseudo-anomalies provide reliable performance estimates.

\subsection{Which pseudo-anomalies are informative?}

Metadata-based pseudo-anomalies generally achieve the best performance, particularly Diverse and Cross-Attribute, while Cross-Class is also competitive. Using real samples from other domains, classes, or attributes, these constructions yield informative pseudo-outlier scores that may better approximate real anomaly scores than those from synthetic perturbations, even in the embedding space.
This is consistent with previous work in \ac{asd}, where related machine types or attributes are used as auxiliary supervision during representation learning \citep{wilkinghoff2024why}.

AutoUAD does not perform better than the inlier variance estimator (-0.14 pp [-0.89, 0.66]), suggesting that its Gaussian pseudo-anomalies provide limited information for model selection in our setting. In contrast, Feature pseudo-anomalies provide a highly competitive, metadata-free alternative. Despite their simpler construction, they achieve performance comparable to metadata-based approaches and outperform AutoUAD (+0.70 pp [+0.11, +1.29]). Their performance is particularly strong with \ac{ldn}, where the AUC Bound even surpasses the Fixed Selection baseline on evaluation data (+1.09 pp [+0.05, +2.09]). We therefore recommend Feature pseudo-anomalies as the default.

Sequence and Element, as alternative metadata-free constructions, consistently perform poorly, while Random pseudo-anomalies also yield weak performance with Pseudo AUC. The AUC Bound is considerably more robust to Random pseudo-anomalies, supporting the use of moment-based estimation when pseudo-anomalies are uninformative.
\Cref{appendix:non_positive} further examines these observations by quantifying the frequency of non-positive mean score differences and perfect pseudo-AUC values across constructions.

Cross-Domain pseudo-anomalies behave differently. Despite representing realistic domain shifts, they are generally less effective with \ac{ldn} than with conventional \ac{knn} distances. As discussed in \cref{sec:ldn_analysis}, \ac{ldn} suppresses domain-mismatch effects in the score, making purely domain-induced deviations less informative. Consequently, semantically richer constructions such as Cross-Attribute and Diverse remain more effective after normalization.

\subsection{Should the full pseudo-outlier score distribution be modeled?}

The AUC Bound generally provides more reliable model selection than Pseudo AUC, achieving lower regret for most pseudo-anomaly constructions. Although Pseudo AUC achieves slightly better candidate rankings with Diverse pseudo-anomalies, this does not translate into a meaningful improvement in model selection. Overall, the AUC Bound is more robust across pseudo-anomaly constructions. This is particularly relevant for less informative pseudo-anomalies, where Pseudo AUC often equals 1 for multiple candidate systems, making selection difficult (see \cref{appendix:non_positive}).

We also examined the domain weighting introduced in \cref{sec:domain_shift}. Using the empirical mixture weight $\lambda=0.99$ performs similarly to using the full reference set, whereas equal weighting ($\lambda=0.5$) substantially degrades performance. This is likely due to the small number of target-domain samples (10 per task), which makes their score moments difficult to estimate reliably. Detailed results are provided in \cref{appendix:mixture_modeling}.

\subsection{How does \acs{ldn} affect detection performance and ranking reliability?}
\label{sec:ldn_improvement}

The difference in selected AUC between \ac{knn} and \ac{ldn} reflects both changes in the underlying detection performance and changes in model-selection quality. We disentangle these effects by decomposing it into an oracle contribution and a selection contribution. Specifically, we define
\[
\begin{aligned}
\Delta \mathrm{AUC}_{\mathrm{oracle}}
&= \mathrm{AUC}_{\mathrm{oracle}}^{\mathrm{LDN}}
 - \mathrm{AUC}_{\mathrm{oracle}}^{\mathrm{KNN}},\\
\Delta R
&= R^{\mathrm{KNN}}-R^{\mathrm{LDN}},
\end{aligned}
\]
where $R$ denotes the regret relative to the best candidate. The selected AUC difference satisfies
\[
\Delta \mathrm{AUC}_{\mathrm{selected}}
=
\Delta \mathrm{AUC}_{\mathrm{oracle}}+\Delta R.
\]
The oracle contribution is estimator-independent, as it depends only on the candidate systems.

\begin{table*}[t]
\centering
\caption{
Decomposition of the selected-system performance improvement obtained with \ac{ldn} relative to \ac{knn}. Entries report paired AUC differences (in percentage points) with 95\% bootstrap confidence intervals. Positive values indicate improvements with \ac{ldn}.
}
\label{tab:ldn_improvement}

\setlength{\tabcolsep}{5pt}
\begin{adjustbox}{max width=\textwidth}
\begin{NiceTabular}{l*{6}{c}}
\toprule

&
\multicolumn{3}{c}{Development}
&
\multicolumn{3}{c}{Evaluation}
\\

\cmidrule(lr){2-4}
\cmidrule(lr){5-7}

Estimator
&
$\Delta \mathrm{AUC}_{\mathrm{oracle}}\uparrow$
&
$\Delta R\uparrow$
&
$\Delta \mathrm{AUC}_{\mathrm{selected}}\uparrow$
&
$\Delta \mathrm{AUC}_{\mathrm{oracle}}\uparrow$
&
$\Delta R\uparrow$
&
$\Delta \mathrm{AUC}_{\mathrm{selected}}\uparrow$
\\

\midrule

Inlier Variance
&
$+1.21\;[-0.06, +2.36]$
&
$+0.49\;[-0.85, +1.90]$
&
$+1.70\;[+0.03, +3.48]$
&
$+2.76\;[+1.61, +3.96]$
&
$+0.11\;[-1.03, +1.24]$
&
$+2.87\;[+1.51, +4.21]$
\\

AUC Bound (Feature)
&
$+1.21\;[-0.06, +2.36]$
&
$+0.93\;[-0.89, +2.73]$
&
$+2.13\;[+0.23, +4.19]$
&
$+2.76\;[+1.61, +3.96]$
&
$+1.50\;[+0.16, +2.93]$
&
$+4.26\;[+2.64, +6.02]$
\\

\bottomrule
\end{NiceTabular}
\end{adjustbox}

\end{table*}

\Cref{tab:ldn_improvement} shows that \ac{ldn} improves oracle detection performance on both development and evaluation tasks, with a statistically significant gain on evaluation. These performance gains are consistent with the analysis in \cref{sec:ldn_analysis}, which predicts that removing leading-order density effects reduces score variability and between-domain score mismatch. Interestingly, \ac{ldn} also improves model-selection reliability, although to a lesser extent. The reduction in regret is small for the inlier-variance estimator, which is not surprising given that inlier variance alone is a weak, representation-dependent proxy for detection performance. The feature-based AUC Bound benefits more, with a statistically significant reduction in regret on evaluation. Together, the improvements in detection performance and model selection yield significantly higher selected AUC for both estimators on both datasets.

We also evaluated whether the estimators can identify which scoring function (\ac{knn} or \ac{ldn}) achieves higher AUC for a given candidate--split pair. Inlier variance identified the better-performing scoring function in 61.4\% [51.4, 71.0] / 72.1\% [62.6, 80.7] of cases on development/evaluation data. However, these rates are no better than simply choosing \ac{ldn} every time: \ac{ldn} has lower inlier variance for every candidate--split pair. The feature-based AUC Bound achieved agreement rates of 64.5\% [55.8, 72.3] / 66.2\% [58.3, 73.7], but did not significantly outperform always choosing \ac{ldn} on development data (+3.01 pp [-5.76, +12.13]) and performed significantly worse on evaluation data (-5.89 pp [-12.32, -0.10]). Thus, neither estimator provides a reliable basis for choosing between the two scoring functions. Given the theoretical motivation for \ac{ldn} and its higher overall detection performance in our experiments, we recommend using it as the default scoring function rather than attempting to select between \ac{knn} and \ac{ldn} based on these estimators.

\section{Conclusion}
\label{sec:conclusion}

In this work, we investigated whether the performance of \ac{knn}-based anomalous sound detectors can be estimated without anomalous validation data. We derived an AUC lower bound in terms of score means and variances, and used local scaling to relate inlier score variability to local geometry of the data. Experimental results show that embedding-space geometry contains useful information about detector performance, but inlier score variance alone is insufficient for comparing different representations. We therefore complemented the geometric analysis with pseudo-anomaly-based estimates of score separation. Across the DCASE 2022–2025 benchmarks and 208 candidate systems, pseudo-anomaly-based estimation substantially improves model selection over inlier variance alone. The AUC Bound provides more robust selection across pseudo-anomaly constructions, while Diverse pseudo-anomalies combined with Pseudo AUC achieve the lowest overall regret. Importantly, our simple metadata-free Feature construction outperforms the Gaussian pseudo-anomaly baseline of AutoUAD, without requiring a generative model or semantic metadata. Combined with \ac{ldn}, the Feature-based AUC Bound even surpasses conventional development-set selection on evaluation data. We therefore recommend Feature pseudo-anomalies as a practical default, while metadata-based constructions can provide further improvements in some settings.

Future work should investigate more informative pseudo-anomaly constructions without semantic metadata and extend the approach to hyperparameter optimization and representation learning. While our experiments focus on acoustic embeddings, the approach is applicable to \ac{knn}-based anomaly detection in other modalities. Its extension to other scoring paradigms remains an interesting direction for future work.

\subsection*{AI use statement}

We used generative AI tools to search for relevant literature, improve the language and readability of the manuscript, create the plot for Figure 1, and obtain feedback on the manuscript using the Google Paper Assistant Tool (PAT). We reviewed all AI-assisted work and take responsibility for the final content of the paper.

\subsection*{Reproducibility Statement}

The pseudo-anomaly constructions are described in \cref{sec:pseudo_anomalies}, while \cref{sec:setup} provides the dataset, candidate-system, and evaluation details. Additional theoretical analysis and embedding-model configurations are provided in \cref{sec:appendix}.

\bibliography{refs}
\bibliographystyle{iclr2027_conference}

\appendix
\section{Appendix}
\label{sec:appendix}

\subsection{Dependence-Robust AUC Bound}
\label{sec:auc_bound_proof}

Even without independence, the AUC can be bounded using only the marginal score variances. Under conditional independence, this bound recovers \cref{eq:auc_variance_bound}.

\begin{proof}
Applying Cantelli's inequality \citep{ion2023sharp} to $\Delta$ gives
\[
\mathbb{P}(\Delta \le 0)
\le
\frac{\operatorname{Var}(\Delta)}
{\operatorname{Var}(\Delta)+\mu_\Delta^2}.
\]
Hence,
\[
\mathrm{AUC}
=
1-\mathbb{P}(\Delta\le0)
\ge
\frac{\mu_\Delta^2}
{\operatorname{Var}(\Delta)+\mu_\Delta^2}.
\]

Expanding the variance gives
\[
\operatorname{Var}(\Delta)
=
\operatorname{Var}(z_{\mathrm{out}})
+
\operatorname{Var}(z_{\mathrm{in}})
-
2\operatorname{Cov}(z_{\mathrm{out}},z_{\mathrm{in}}).
\]
By Cauchy--Schwarz,
\[
\left|\operatorname{Cov}(z_{\mathrm{out}},z_{\mathrm{in}})\right|
\le
\sqrt{
\operatorname{Var}(z_{\mathrm{out}})
\operatorname{Var}(z_{\mathrm{in}})
},
\]
and therefore
\[
\operatorname{Var}(\Delta)
\le
\left(
\sqrt{\operatorname{Var}(z_{\mathrm{out}})}
+
\sqrt{\operatorname{Var}(z_{\mathrm{in}})}
\right)^2.
\]

Combining these inequalities yields
\[
\mathrm{AUC}
\ge
\frac{\mu_\Delta^2}
{
\left(
\sqrt{\operatorname{Var}(z_{\mathrm{out}})}
+
\sqrt{\operatorname{Var}(z_{\mathrm{in}})}
\right)^2
+
\mu_\Delta^2
}.
\]

This bound is looser than \cref{eq:auc_variance_bound}, which exploits conditional independence to remove the covariance term.
Under conditional independence of $z_{\mathrm{in}}$ and $z_{\mathrm{out}}$ given
the fixed reference set, the covariance term vanishes, giving
\[
\operatorname{Var}(\Delta)
=
\operatorname{Var}(z_{\mathrm{out}})
+
\operatorname{Var}(z_{\mathrm{in}}),
\]
and hence \cref{eq:auc_variance_bound}.
\end{proof}

\subsection{Embedding Model Configurations}
\label{appendix:model_configs}

All embedding models are used in their publicly available pre-trained form without additional fine-tuning. OpenL3 \citep{cramer2019look} uses the environmental checkpoint with 512-dimensional embeddings extracted from 1\,s windows with a hop size of 0.1\,s. For BEATs \citep{chen2023beats}, we use the official Iter3 checkpoint pre-trained on AudioSet \citep{gemmeke2017audioset}. For EAT \citep{chen2024eat}, we use the official large checkpoint pre-trained for 20 epochs on AudioSet, while Dasheng \citep{dinkel2024dasheng} uses the official base checkpoint.


\subsection{Frequency of non-positive mean score differences}
\begin{table*}[t]
\centering
\caption{
Frequency of non-positive mean score differences and perfect
pseudo-AUC values for different pseudo-anomaly constructions.
Entries report the empirical percentages of candidate--split
observations with $\Delta\mu\leq 0$ and
$\mathrm{AUC}_{\mathrm{pseudo}}=1$, respectively.
}
\label{tab:non_positive_mean_difference}

\setlength{\tabcolsep}{3pt}
\begin{adjustbox}{max width=\textwidth}
\begin{NiceTabular}{l*{8}{c}}
\toprule

&
\multicolumn{4}{c}{\ac{knn}}
&
\multicolumn{4}{c}{\ac{ldn}}
\\

\cmidrule(lr){2-5}
\cmidrule(lr){6-9}

&
\multicolumn{2}{c}{Development}
&
\multicolumn{2}{c}{Evaluation}
&
\multicolumn{2}{c}{Development}
&
\multicolumn{2}{c}{Evaluation}
\\

\cmidrule(lr){2-3}
\cmidrule(lr){4-5}
\cmidrule(lr){6-7}
\cmidrule(lr){8-9}

Pseudo-Anomaly
&
$\Delta\mu\leq 0$
&
$\mathrm{AUC}_{\mathrm{pseudo}}=1$
&
$\Delta\mu\leq 0$
&
$\mathrm{AUC}_{\mathrm{pseudo}}=1$
&
$\Delta\mu\leq 0$
&
$\mathrm{AUC}_{\mathrm{pseudo}}=1$
&
$\Delta\mu\leq 0$
&
$\mathrm{AUC}_{\mathrm{pseudo}}=1$
\\

\midrule

Random
& 0.00 & 100.00 & 0.00 & 99.58
& 0.00 & 96.51 & 0.00 & 97.04
\\

Sequence
& 3.82 & 6.12 & 9.84 & 2.03
& 29.36 & 3.66 & 31.74 & 0.84
\\

Feature
& 0.05 & 24.87 & 0.02 & 18.76
& 0.00 & 18.96 & 0.00 & 16.24
\\

Element
& 4.05 & 16.51 & 7.96 & 12.16
& 23.86 & 12.72 & 25.61 & 9.71
\\

Cross-Domain
& 0.00 & 6.86 & 0.00 & 14.27
& 0.00 & 0.53 & 0.00 & 8.95
\\

Cross-Class
& 0.00 & 14.43 & 0.00 & 37.46
& 0.00 & 10.68 & 0.00 & 33.48
\\

Cross-Attribute
& 0.00 & 1.14 & 0.00 & 4.48
& 0.00 & 0.00 & 0.00 & 3.03
\\

Diverse
& 0.00 & 0.55 & 0.00 & 1.60
& 0.00 & 0.11 & 0.00 & 1.34
\\

\bottomrule
\end{NiceTabular}
\end{adjustbox}
\end{table*}
\label{appendix:non_positive}

In \cref{tab:non_positive_mean_difference}, we investigate whether the assumption of a positive mean score difference is violated. This occurs predominantly for pseudo-anomaly constructions that yield poor separation from inliers, such as Sequence and Element. Most constructions also frequently yield perfect pseudo-AUC values, limiting their ability to discriminate between candidate systems and explaining their poor performance as pseudo-AUC-based selection criteria. In contrast, the AUC Bound only approaches 1 when the score variances of both inliers and pseudo-outliers vanish (for nonzero mean separation) or when the mean separation tends to infinity.

\subsection{Does domain-mixture modeling improve ranking reliability?}
\label{appendix:mixture_modeling}

\begin{table*}[t]
\centering
\caption{
Domain-specific effect of domain weighting on Feature-based AUC-bound selection. Entries report Spearman rank correlation ($\rho_{\mathrm{rank}}$), pairwise ranking accuracy ($A_{\mathrm{pair}}$), and regret ($R$, in percent). The mixture weight $\lambda=0.99$ corresponds to the empirical domain proportions (i.e., the full reference set), whereas $\lambda=0.5$ assigns equal weight to both domains.
}
\label{tab:domain_weighting}

\setlength{\tabcolsep}{2.5pt}
\begin{adjustbox}{max width=\textwidth}
\begin{NiceTabular}{l*{19}{c}}
\toprule

&
&
\multicolumn{6}{c}{Overall}
&
\multicolumn{6}{c}{Source Domain}
&
\multicolumn{6}{c}{Target Domain}
\\

\cmidrule(lr){3-8}
\cmidrule(lr){9-14}
\cmidrule(lr){15-20}

&
&
\multicolumn{3}{c}{Development}
&
\multicolumn{3}{c}{Evaluation}
&
\multicolumn{3}{c}{Development}
&
\multicolumn{3}{c}{Evaluation}
&
\multicolumn{3}{c}{Development}
&
\multicolumn{3}{c}{Evaluation}
\\

\cmidrule(lr){3-5}
\cmidrule(lr){6-8}
\cmidrule(lr){9-11}
\cmidrule(lr){12-14}
\cmidrule(lr){15-17}
\cmidrule(lr){18-20}

Detector
&
$\lambda$
&
$\rho_{\mathrm{rank}}\uparrow$
&
$A_{\mathrm{pair}}\uparrow$
&
$R\downarrow$
&
$\rho_{\mathrm{rank}}\uparrow$
&
$A_{\mathrm{pair}}\uparrow$
&
$R\downarrow$
&
$\rho_{\mathrm{rank}}\uparrow$
&
$A_{\mathrm{pair}}\uparrow$
&
$R\downarrow$
&
$\rho_{\mathrm{rank}}\uparrow$
&
$A_{\mathrm{pair}}\uparrow$
&
$R\downarrow$
&
$\rho_{\mathrm{rank}}\uparrow$
&
$A_{\mathrm{pair}}\uparrow$
&
$R\downarrow$
&
$\rho_{\mathrm{rank}}\uparrow$
&
$A_{\mathrm{pair}}\uparrow$
&
$R\downarrow$
\\

\midrule

\multirow{2}{*}{\ac{knn}}
& $0.99$
& 0.22 & 0.59 & 6.00 & 0.36 & 0.64 & 5.06
& 0.28 & 0.61 & 6.53 & 0.38 & 0.65 & 6.28
& 0.11 & 0.55 & 9.17 & 0.26 & 0.60 & 7.25
\\
& $0.50$
& 0.14 & 0.55 & 7.27 & 0.25 & 0.59 & 6.17
& 0.16 & 0.56 & 8.25 & 0.24 & 0.59 & 8.40
& 0.05 & 0.52 & 9.88 & 0.15 & 0.56 & 8.23
\\

\midrule

\multirow{2}{*}{\ac{ldn}}
& $0.99$
& 0.23 & 0.58 & 5.49 & 0.41 & 0.65 & 4.28
& 0.25 & 0.59 & 6.66 & 0.38 & 0.64 & 5.40
& 0.19 & 0.57 & 6.47 & 0.32 & 0.61 & 6.53
\\
& $0.50$
& 0.14 & 0.55 & 8.15 & 0.25 & 0.59 & 5.48
& 0.14 & 0.55 & 8.42 & 0.26 & 0.59 & 6.28
& 0.14 & 0.55 & 7.89 & 0.17 & 0.56 & 8.48
\\

\bottomrule
\end{NiceTabular}
\end{adjustbox}

\end{table*}

We also examined the effect of domain weighting on model selection. With the empirical mixture weight ($\lambda=0.99$), \ac{ldn} achieves lower target-domain regret than \ac{knn}, reducing it from 9.17\% to 6.47\% on development data and from 7.25\% to 6.53\% on evaluation data. However, assigning equal weight to both domains ($\lambda=0.5$) consistently worsens ranking performance and increases regret, both overall and within each domain. This is likely because only 10 of the 1,000 reference samples come from the target domain, making its score-moment estimates unreliable. The test sets, in contrast, contain equal numbers of source- and target-domain samples. Thus, although the domain-wise analysis helps explain the effects of domain shift and \ac{ldn}, explicitly balancing the domains during model selection offers no practical benefit in our setting, and using the full reference set with its empirical proportions is preferable.

\end{document}